# AI SAFETY: A CLIMB TO ARMAGEDDON?

Herman Cappelen (*University of Hong Kong*), Josh Dever (*University of Texas at Austin*) and John Hawthorne (*Australian Catholic University, University of Southern California*)

AI is regularly described as an existential threat to humanity, on a par with nuclear and biological weapons.[1] The worry is this: if we create an AI that has super-human capacities, it could use those capacities to achieve any kind of goal. Some of those goals might require destroying humanity. This could be intentional or accidental, a result of misalignment or of misuse. Either way, the argument goes, super intelligent AI can easily result in an Armageddon.

In response to this and related worries, there's an effort to make AI safe, which is often combined with an effort to align the values (or goals) of AI with ours.[2] We should try to create AI that protects, and ideally promotes, human welfare.

In this paper, we present a surprising and counterintuitive argument against AI safety. Rather than denying the existential risk posed by AI, our argument starts from the assumption that AI does indeed pose an existential threat to humanity. We then show that, under certain assumptions, safety measures are not merely pointless or useless, but actively dangerous. The implications of this argument are potentially far-reaching. If the argument is sound, it suggests that the very efforts we are undertaking to mitigate the risks of AI may, in fact, be increasing those risks.[3]

---

[1] The literature here is vast and growing. The 2023 Stanford University Artificial Intelligence Index Report says that a survey of AI researchers shows that 36% of them agree or weakly agree that in the next century AI could cause a disaster at the level of a full-scale nuclear war. Bostrom (2014) is a locus classicus, expanding on influential earlier papers such as his (2012) which introduced terminology like the orthogonality thesis and instrumental convergence (neither of which, as will be evident, bear on our discussion). Ord 2020 explicitly considers AI alongside nuclear and biological weapons in a survey of existential risk as does, albeit at less length, the influential MacAskill 2022. For a general overview see Brungage et al 2018.

[2] See Russell 2019; Christian 2020; see also the discussion of 'friendly AI' in Yudkowsky 2008.

[3] Others have related concerns. For example, Hendryks says, "'Often, though not always, safety and general capabilities are hard to disentangle. Because of this interdependence, it is important for researchers aiming to make AI systems safer to carefully avoid increasing risks from more powerful AI systems."

At the core of much interesting philosophy, we find apparently sound reasoning that leads to a highly implausible conclusion. That's the spirit in which to take our anti-safety argument. By grappling with it, we are forced to re-examine our fundamental assumptions about AI safety. In the last part of the paper, we explore different strategies for responding to the anti-safety argument. It turns out that the argument is remarkably robust and that even when its central assumptions are weakened, it holds up. We find this surprising, challenging, and also somewhat disturbing.

Here is the structure of what follows: We first present a motivating example, The Doomed Rock Climber, and then use that to develop an anti-AI safety argument (the Non-Deterministic Argument). In the last part of the paper we consider three reply strategies: Optimism, Mitigation, and Holism. Our responses to these focus on what we call: Bottlenecking, Perfection Barrier, and Equilibrium Fluctuation.

## 1. Motivating Example: The Doomed Rock Climber

Here is a case where we think safety measures are, indisputably, dangerous. This case, we'll argue, instantiates a pattern that we can also find in AI:

> *The Doomed Rock Climber*: You're watching a rock climber start a climb that you are completely convinced she will be unable to complete. Our credence that she will fall is 1. When she's 4 feet above ground, she asks you for help. She wants some chalk to help her get higher. The chalk is a safety device: it will help her climb higher. It will not prevent her from falling. So the effect of the added safety is to help her climb to a point where failure is worse than at 4 feet, and potentially catastrophic (i.e. results in death.)

It would be irresponsible to provide safety measures (i.e. chalk). Increased safety would allow the climber to keep getting higher. Since she'll eventually fall, it is better for her to fall at an earlier stage where she just breaks a few bones. A fall from higher up is an existential threat and giving her tools to enable more climbing is wrong.

Suppose our attitude towards the climber is as follows::

1. We are certain that the climber will fall.
2. We are sure that the higher the fall, the worse things will be[4].
3. We are sure that providing chalk to the climber will result in a higher fall.

In this situation it is clear that providing safety (i.e. chalk) to the climber is a bad idea. These observations can easily be played out in a simple decision theory: under these constraints, the expected height of fall will be higher conditional on providing chalk than on not providing the chalk and so, given that we are sure that badness correlates with height of fall, the expected utility of providing chalk will be lower than the expected utility of not providing chalk.

The structure can be instantiated in all sorts of ways. Here is one instantiation that is relevant to the theme of this paper:

> An AI system is in operation and is becoming increasingly more powerful. We are sure the system will suffer a security failure. We are sure that the more powerful the system is at the point at which a safety failure occurs, the worse things will be. We have the option of instituting a safety measure and are certain that instituting it will make it such that the first safety failure occurs when the system is in a more powerful state than it would be, were we not to provide the safety.

Again, if all the assumptions made here are true, we should not provide safety. Instituting the safety measure has lower expected utility than not instituting them.

This structure can be generalized. Let F be a kind of failure, P a kind of parameter, and S a kind of safety measure, one that can come with various degrees of badness. Let say that a situation is a **Simple Anti-Safety Situation** with respect to F, P and S just so long as the following obtains:

> (i) **Certain Failure:** We are sure that an F will occur.

---

[4] Maybe this isn't completely realistic in the climber case because when the climber gets to a height that is fatal, the fall won't get any worse the higher she climbs. It is more realistic in the AI case, discussed below, because there it is not clear that there's a maximum damage (the AI could kill us all, but it could also enslave and torture us for eternity etc.). If there's a max damage level, then the argument applies until we get to a height that might be fatal.

(ii) **Certain Increasing Badness**: We are sure that the higher the value of P at which the first F occurs, the worse things will be.

(iii) **Certain Assistance**: We are sure that providing S will make for a higher value of P for the first F.

Then providing F will be a bad idea: not providing it will have higher expected utility than providing it. Moreover, assuming you should do that which has the highest expected utility, one should not provide F.

## 2. A Non-Deterministic Version of the Argument

Assumptions 1-3 above might seem strong. They assume certainty about future failure, increased badness, and about the effects of assistance. While we think those assumptions can be defended, we don't think doing so is necessary because a version of the argument above goes through even if you don't have complete confidence in the core assumptions. Consider again the Doomed Climber: maybe we can't be quite sure how bad a fall from a certain height will be. Suppose there are spikes and pillows on the ground and the damage done by a fall depends not just on the height of the fall but on whether the climber falls on the spikes or the pillows[5]. Applied to AI safety, maybe a more powerful AI system will also help us create increasingly powerful safety devices and so failure by a more powerful AI won't necessarily be worse than failure by a less powerful AI. These and other natural uncertainties do not block the kind of anti-safety argument we have introduced. Even given pillows and spikes on the ground, it may be that the *expected* damage from a high fall is higher than the *expected* damage from a low fall. To accommodate these less deterministic cases, we will in what follows work with a non-deterministic version of the argument we gave earlier:

1. **Certain Failure[6] Premise:**

---

[5] Another option: maybe the badness of the world depends on chance situations *other than the fall*: Maybe a high fall will inspire the climber's mother to save hundreds of other climbers in a way that the low fall will not.

[6] If you want more precision about what we mean by 'failure', we can give you this: 'failure' in the AI case is an event that's directly responsible for the death of 10,000 people or more. For the climber: a fall all the

a. **Cliff version:** we are sure that a fall will occur
   b. **AI version**: we are sure that a safety failure will occur.

2. **Getting Worse Premise**:

   a. **Cliff Version:** the higher you expect the climber to be when she first falls, the worse you expect the world to be.[7]
   b. **AI version**: the more powerful you expect AI to be when it first fails, the worse you expect the world to be.

3. **Comparison Premise**.

   - **Cliff version:** the expected height of the first fall is higher if you provide safety than if you don't.
   - **AI version**: the expected level of power when the AI first fails is higher if you provide safety than if you don't.

4. **Common decency Premise**: if you expect things to go worse if you do something than if you don't do that thing, then you shouldn't do it[8].

The non-deterministic version of the argument acknowledges uncertainty by introducing probabilities. Instead of assuming that providing chalk will definitely lead to a higher fall, we consider the probability that it will do so. Similarly, instead of assuming that a higher fall will definitely lead to worse outcomes, we consider the expected value of the outcomes, taking into account the probabilities of different scenarios. Note that we are still assuming that a failure will occur (Certain Failure Premise). Below, we'll look at various ways to reject that premise as well. What we want to notice at this point is simply that the four premises of the non-deterministic

---

way to the ground that results in death. The argument will work for various other definitions of 'failure' as well, but the choice is delicate.

[7] Careful version -- the ranking of actions by expected height of fall matches the ranking of actions by expected badness (and is the inverse of the ranking of actions by expected goodness).

[8] What if you've promised to do something, but expect that things will get worse if you keep your promise? We recognize that there are moral philosophers who think the Common Decency premise is false in its full generality, but we think it's a bit desperate to think that is the key escape route to the argument in this paper.

argument *still imply that we should not provide chalk to the climber and should not provide AI safety.* If you find that conclusion repugnant in the case of AI, you are faced with a puzzle: you need to decide which premise to give up.

Note that our argument doesn't target all safety measures - it's not, for example, an argument against sterilizing surgical instruments, implementing checklists for airplane maintenance, locking your door when you leave your house, or putting names on medications in hospitals. The key difference lies in the nature of the risks and the potential for harm in these cases compared to the challenges posed by AI systems. The safety measures for surgical instruments don't obviously enable the instruments to become more dangerous over time. Airplane maintenance checklists don't obviously increase the potential for harm in the future[9]. Locking your door does not help develop more powerful or dangerous burglaries. In short, the Getting Worse and Comparison Premises in the non-deterministic version of the argument don't easily generalize[10].

## 3. Responses: Optimism, Mitigation, and Holism

There are many strategies for responding to the anti-safety argument. As a heuristic, we'll classify replies into three categories[11]:

1.   **Optimism:** optimists think that we'll keep getting better at safety and we'll manage to stay ahead of the dangers posed by AI. In its most natural version this involves faith in human expertise, but it might also involve optimism about AI's willingness to protect us (for example, one might be hopeful that AI will recognize when AI is tending toward Armageddon and then destroy computer technology and commit a form of suicide.)

---

[9] In cases where it does lead to more danger, the downsides are obviously outweighed by the benefits: maybe checklists for airplanes contributed to larger airplanes, which resulted in more damaging crashes. However, there is broad agreement that the benefits of global mass transportation considerably outweigh the cost of the occasional crash. If you don't agree with that consensus, you in effect agree with our argument. Moreover, if you thought we were moving in the direction of transportation devices that could ship millions of people in one vehicle, you might be more troubled. Then our argument will indeed become more forceful when applied to transportation as well.

[10] There's an interesting question about the extent to which the argument applies to nuclear safety measures. Initially, it seems that our argument does not generalize because the Getting Worse and Comparison premises don't apply. If, however, there are knock-on effects, where one accident triggers additional nuclear events, that changes the calculation. We'll leave as an open question the extent to which the arguments here apply to this domain.

[11] Of course there are many mixed responses that combine elements of (or non-negligible credence in) the three responses.

2.     **Holism:** this is a broad cluster of replies that have in common that they agree with the non-deterministic argument that safety measures will make the first failure worse. However, the Holists argue that this is only true locally, and that our focus should be on the global consequences of our alternative courses of action. There's a wide array of possible forms of Holism, and below we focus on three of them: Learning-from-Failure-Holist, Net-Positive-Impact-Holistic, and Point-of-No-Return Holist.

3.     **Mitigation:** Mitigators advocate for measures that aim to reduce the damage caused when a fall inevitably occurs. This contrasts with a focus on safety measures that prevent the climber from falling or prevent AI failure. In the case of the Doomed Climber, we could install safety nets to catch the climber or place cushioning materials at the base of the climb to soften the impact of a fall. In the context of AI safety, this could involve techniques designed to limit the scope and severity of an AI malaction.

These three strategies - Optimism, Holism, and Mitigation - target different premises of the non-deterministic argument. The Optimist rejects the Certain Failure Premise, arguing that it is possible for safety measures to improve at a faster rate than the increase in danger, thereby preventing a guaranteed safety failure. The Mitigator, on the other hand, accepts the Certain Failure Premise but rejects the Getting Worse Premise. It suggests that even if a safety failure is inevitable, the severity of the outcome can be reduced by implementing measures that limit the scope and impact of the failure. The Holist challenges the Comparison Premise, arguing that focusing on individual incidents fails to account for the complex, interconnected nature of AI systems. This strategy suggests that the benefits of safety measures may manifest in ways that are not captured by the simplified comparison of expected outcomes in the non-deterministic argument.

An initial lesson from this paper is that those who advocate for various kinds of AI-safety measures should be clear on which (combination) of strategies they endorse. The next step is to justify the strategy. What we aim to show in the rest of this paper is that all three responses face serious challenges. The challenges stem from three features of AI that we call: **Bottlenecking**, **Perfection Barrier**, and **Equilibrium Fluctuation**.

## 3.1. On Optimism

What we have in mind here is not 'simple optimism', i.e. the view that our safety measures are good enough to cut down radically on the number of disasters. This is what we normally mean by safety optimism, but it is non-responsive to our argument. Cutting down on the number of disasters entails increasing expected time to the next disaster and that entails making the next disaster worse. That's exactly what the argument points out is bad.

The interesting kind of optimism is Optimistic Optimism. This is the idea that safety can improve faster than danger is increasing, leading to a decrease in the expected badness over time. To see clearly what's required from Optimistic Optimism, we need to be clear on possible increases in danger from AI systems:

- *High Danger:* as AI systems become more powerful, the magnitude of potential harm reaches extremely high levels.
- *Arbitrary Danger:* in the long run, as AI systems continue to advance, the potential for harm may become arbitrarily large: there is no inherent limit to the damage that could be caused.

If some versions High Danger and Arbitrary Danger are true, the rate at which safety measures improve must not only keep up with the increasing risks but must actually surpass them in terms of the speed of improvement. It is not enough for safety to improve incrementally; it must achieve a level of robustness that can effectively counter the immense potential for damage. To address Arbitrary Danger, safety measures must become arbitrarily close to perfect, approaching a level of infallibility that can effectively prevent harm even in the face of unbounded risk.

### 3.1.1. Against Optimism

There are three features of AI that we think make it unlikely that we'll ever develop sufficiently extraordinary safety measures: Bottlenecking, Perfection Barrier, and Equilibrium Fluctuation.

- **Bottlenecking**: Safety mechanisms must route through fallible bottlenecks -- human implementers, regulators setting rules, people programming safety code, physical devices to shut down power, etc. There's a realistic hard cap on how high reliability can get for these bottlenecks. Arbitrarily high is too much to ask. Even very high can be hard to achieve.

- **Perfection Barrier**: Incredibly good safety aims at a kind of perfection -- we need to be absolutely sure that things of a certain sort never happen. Perfection gets asymptotically harder to achieve because it requires very fine control. Getting the dart to land in the bullseye that occupies 10% of the target can be more than 10 times as hard as getting it to hit the target, getting it to land in the inner 10% of the bullseye can again be more than 10 times as hard. This contrasts with the increasing damage potential: *damage doesn't have to be finely tuned*. Just the random exertion of powerful causal capacities will do lots of damage. This makes it incredibly hard for safety to keep up.

- **Equilibrium fluctuation**: In response to the challenges posed by the bottlenecks and the perfection barrier, the Optimistic Optimist might suggest leveraging the AI's increasing causal capacities to develop and implement safety measures that keep pace with the AI's growing potential for harm. Think of Equilibrium Fluctuation as a situation where the AI's capabilities to develop and implement safety measures are evenly matched with its potential to cause harm, leading to random fluctuations in the relative effectiveness of safety and harm over time.

This might seem like an initially promising thought, but note that the same advanced capabilities that could be used for safety could just as easily be used to cause harm, either intentionally or unintentionally. Due to the perfection barrier, achieving near-perfect safety is much more difficult than causing harm. Safety requires precise, targeted interventions, while harm can often result from relatively simple, untargeted actions. This asymmetry implies that even if the AI's capabilities are being leveraged for both safety and harm, the potential for harm may still outstrip the potential for safety in the long run.

Moreover, even if the AI's safety and harm capacities remain evenly balanced as they grow, random fluctuations in their relative effectiveness over time can lead to catastrophic consequences. During periods when the potential for harm temporarily surpasses the potential for safety, the high stakes and increasing potential for damage mean that even brief imbalances could result in devastating outcomes. As AI systems become more advanced, the longer and more frequent these periods of imbalance become, the greater the cumulative risk posed by the fluctuations. (By analogy, suppose Superman and Lex Luthor are both getting more powerful over time. Even if Superman is generally more powerful, a temporary imbalance in the other direction will induce disaster and the later the first imbalance, the greater the disaster.) The core idea here is that the inherent unpredictability of the AI's development, coupled with the asymmetric challenges posed by the perfection barrier, suggests that this strategy is unlikely to provide a reliable response to our anti-safety argument.[12]

## 3.2. On Holism

The general thought here is that it's a mistake to focus on just one climb by the climber, or just one disaster by the AI. Here are three distinct forms of holism (that can be also be combined):

- **Learning from failure Holism**: by experiencing a more significant first failure, we may gain valuable insights and knowledge that can be used to prevent or mitigate future failures. Consider an analogy with the climber: if we don't give her the chalk and she experiences a minor fall with minimal consequences, she may not learn much about the risks and limitations of her approach. However, if she suffers a more serious fall that results in a broken ankle, she will be forced to confront the reality of the risks she faces and adapt her strategy accordingly. Similarly, a more severe initial AI failure (which our anti-safety argument is trying to prevent) may provide valuable information about the system's vulnerabilities and the effectiveness of existing safety measures, enabling researchers to develop more robust solutions for future iterations.

- **Net positive impact Holism:** Even if we don't learn from the first failure, the AI system may still produce sufficiently many positive consequences. Think of this as the view that the benefits of the AI's continued operation outweigh the risks posed by the potential for further catastrophic failures[13]. Suppose our climber knows that the risk of a fatal fall is high, but believes that the potential rewards - the advancement of mountaineering knowledge, the personal achievement, and the inspiration others may draw from the feat - justify the risk. Even if the climber ultimately perishes in the attempt, they may view their effort as worthwhile if it contributes to the greater good of their field and humanity as a whole. Similarly, Net-Positive-Impact Holists argue that the benefits of AI, such as scientific breakthroughs, technological innovation, and solutions to global challenges, are so immense that they outweigh even the risk of an existential catastrophe.

---

[12] We acknowledge the possibility of a more pessimistic version of Optimism: this is a view that acknowledges the risks associated with advanced AI systems but argues that implementing safety measures is rational even if the probability of success is extremely low, as long as there is a chance of avoiding an existential catastrophe. The thought is that a chance at AI utopia is worth the associated risk. We're not convinced, and an analogy with the bullseye might again help clarify: hitting a small bullseye on a dartboard becomes exponentially harder as the size of the bullseye shrinks, while the consequences of missing the bullseye grow increasingly severe if the surrounding area is filled with hazards. In the context of AI, the damage potential is likely to outstrip the expected benefit of aiming for perfection, as our ability to precisely target high-value outcomes will lag behind the AI's increasing capacity for harm.

- **Point of no Return Holist:** Suppose we are sure that a very bad thing will happen, no matter what we do. We also know that there are less bad things that might happen prior to the very bad thing. In that situation, our argument can be undermined by global considerations: a focus on individual incidents fails to account for the complex, interconnected sequence of events. Here's a toy example: There's a terrorist who plans to blow up a village (2000 people), a town (30,000 people), and a city (400,000 people). You can't do anything about the town and the city, but you can save the village. If we introduce safety and as a result save the village, then the first attack will be bigger (town size, rather than village size). If your goal is to minimize the *first* accident, then it makes sense to *not* save the village. That, however, is not what we want to say. In this case, it is a mistake to think that the ranking of actions by the expected badness of the first attack matches the ranking of actions by the expected overall badness of the world (conditional on performing each of those actions). Conditional on saving the village we expect the first terrorist event to kill 30000, but conditional on not saving the village we expect the first terrorist event to kill 2000. So the latter action does better by the lights of the expected value of the first attack. But clearly we expect things to go better overall by saving the village. Applied to AI: if you're convinced that Armageddon will happen anyway (or that some terrible consequences will follow no matter what we do) – i.e., if we're past *the point of no return* - then why shouldn't we save the world from horrors on the way to Armageddon?

---

[13] Prior to AI-Armageddon, AI might, for example, save us from horrific illnesses, nuclear disasters, and help us fight off alien invasions.

### 3.2.1 Against Holism

The Learning Holist agrees with us on the analysis of the single-shot climb/disaster, so the burden is on them to say how the multi-shot setting makes a difference. The obvious first response to the appeal to the Learning Holist is that, if we're right about a climb, then we're right about each climb, just by repeating the reasoning. If providing safety measures for a single climb leads to a worse outcome, then providing safety measures for each subsequent climb will also lead to worse outcomes, as the same reasoning applies to each individual case. The cumulative effect of these worse outcomes across multiple climbs only strengthens our argument against providing safety measures. So, it doesn't make any difference to take the larger perspective[14].

The Learning-Holist, also faces the concerns we raised above. Bottlenecking suggests that the rate of learning and adaptation may be limited by the human and organizational factors involved in implementing safety measures. Even if lessons are learned from one incident, the speed at which those lessons can be applied to other AI systems may be constrained, making it close to impossible to keep pace with the rapid evolution of AI capabilities across multiple incidents. The Perfection Barrier becomes even more daunting when considering multiple AI incidents. As the number of incidents increases, the cumulative probability of a catastrophic failure also grows, even if safety measures are improving incrementally. It seems implausible that the Learning Holist's emphasis on gradual adaptation is sufficient to overcome the challenge of achieving near-perfect safety across a large number of AI systems and incidents. Finally, the points we made in connection with equilibrium fluctuation take on added significance in the context of multiple AI incidents. As safety measures and AI capabilities evolve over time and across different systems, there may be complex interactions and feedback loops that lead to unexpected and potentially destabilizing fluctuations in the overall AI ecosystem.

The Net-Positive-Impact Holist, who argues that the benefits of advanced AI will ultimately outweigh the risks of catastrophic failure, relies on a highly speculative and uncertain calculus. The view is that the good effects that a super powerful AI will bring, outweigh a final Armageddon. To make that view work, you have to have very high confidence in the value of the kinds of good effects that AI will bring and also about the horrors of a final Armageddon. We don't have strong views here, but keep in mind that for every potential long-term positive effect, there are competing potential long-term negative effects. A super powerful AI could cure cancer, but it could also create a horrific virus that it decides to spread. It could make us live longer lives, but it could also make those long lives utterly horrific. Gambling on global overall goodness strikes us as dangerous. Given the vast range of potential outcomes and the difficulty of predicting and controlling the long-term impacts of AI, it is far from clear that the positive effects of AI will necessarily predominate. The potential for AI systems to optimize for narrow objectives in ways that lead to unintended and potentially devastating consequences, even if the initial intentions were positive, underscores the risks of relying on a net-positive impact argument.

---

[14] The Ultra-Optimistic Learning Optimist will respond that one result of the first failure with safety will end all climbing, forever. Whereas the first failure without safety will be too mild and will not result in the end of all climbing. As our label suggests, we think this is somewhat naive, but we don't have a knock down argument.

Our reply to The Point-of-No-Return-Holist, depends on the details of the case. In the motivating example, there is certainty about both a sequence of events (three terrorist attacks) and that stopping the Village destruction will have no effect on the City and Town destructions. In the case of AI, these kinds of assumptions are, at best, questionable. First, it's implausible to assume that we can be completely sure that there will be an AI Armageddon, no matter what. Even if we develop a superintelligent and powerful AI system, it might just ignore us (and even the earth). We might be to it as ants are to us: irrelevant and largely ignored. Maybe a superintelligent AI will focus on other parts of the universe and other AIs. Complete confidence that this won't happen seems unwarranted. If those kinds of scenarios are epistemically possible, it should weaken your confidence in AI-Armageddon. Then the situation becomes more delicate. To go back to our example of the terrorists, suppose you are pretty confident the terrorist will destroy the town, but you are not sure. Suppose also that there is some chance that if the village is blown up, it increases the chances of the town not being blown up (maybe because the village destruction will trigger increased safety measures in the town). Then the calculation of what to do becomes more delicate. The situation will then at least not be as simple as it is in the original terrorist example. That said, we agree that the net impact of AI safety measures in this kind of "point of no return" scenario is a complicated matter that requires careful holistic consideration.

Second, putting aside the objection above and assuming that we can be sure there will be an Armageddon, safety might affect the *timing* of the Armageddon. The initial thought is that if we are sure AI Armageddon will happen, then we should just make the road to Armageddon as accident-free as possible (this is the pro-safety thought). That, however, overlooks the length of the road, i.e., the time to Armageddon. One possibility is that if we save people along the way, that makes the path to Armageddon shorter (i.e., humanity will be obliterated earlier). Here is one way that could happen: If AI seems safe, it will attract additional funding, additional funding will enable faster progress on creating a dangerous AI, and the result is an earlier AI Armageddon. In short, it is not clear, in expectation, that saving people along the way (through introducing safety measures) reduces the overall badness.

## 3.3. Mitigation

First a somewhat concessive observation about Mitigation. We're not denying that some ordinary mitigation measures can be valuable: obviously, if we know an AI system will infect Alice with a potentially lethal virus tomorrow, our argument doesn't count against developing effective antivirals. We have no quarrel with ordinary mitigation efforts (unless they encourage faster development of more powerful AI systems - more on that below).

If, however, you are worried about a final AI Armageddon, Mitigation isn't going to do all that much to defuse your concern and the reasons are familiar from the discussions above:

- **Mitigation and Bottlenecking:** the same bottlenecks that limit the effectiveness of safety mechanisms also constrain our ability to mitigate the risks posed by advanced AI systems. Mitigation measures often rely on human oversight, regulatory frameworks, and physical fail-safes—all of which are subject to the same limitations as the safety mechanisms they are meant to support. Even if we strive for extremely high reliability in these bottlenecks, achieving perfection is likely an unrealistic goal. As AI systems become more advanced and complex, the potential for harm grows exponentially. To keep pace with this increasing damage potential, our mitigation measures would need to achieve an extraordinarily high level of reliability. However, the Bottlenecking effect suggests that there is a fundamental limit to how effective these measures can be. No matter how much effort we put into improving our mitigation strategies, we may never be able to fully overcome the constraints imposed by the fallible bottlenecks they must route through.

- **Mitigation and the Perfection Barrier:** the Perfection Barrier poses an even greater challenge for Mitigation than it does for prevention. In the context of Mitigation, we are confronted with the task of undoing the destructive effects of an AI system that has already begun to cause harm. This is inherently more difficult than preventing the harm from occurring in the first place. Consider an analogy: creating a ripple in a pond is easy; one simply needs to throw a rock. However, undoing that ripple requires incredibly fine-tuned interventions. As the ripple grows larger, the complexity of the intervention needed to counteract it increases dramatically. Some effects, such as the loss of life, may be impossible to mitigate altogether. The same principle applies to AI-induced catastrophes. The destructive effects of an advanced AI system can be essentially random and widespread, while the measures needed to mitigate those effects must be precisely targeted and carefully orchestrated. As the scale of the destruction increases, the difficulty of implementing effective mitigation measures grows exponentially. Moreover, some forms of damage may be fundamentally irreversible. While an AI system might be able to rebuild infrastructure destroyed by a catastrophic event, it cannot bring back lives that have been lost.

- **Mitigation and Equilibrium Fluctuation:** even if we set aside the Perfection Barrier and assume that mitigation and damage potential remain balanced, the increasing scale and complexity of AI systems will inevitably lead to random fluctuations in this equilibrium. As AI systems become more advanced, the range of potential fluctuations between mitigation and damage will increase. A minor lapse in mitigation can result in devastating damage when dealing with more powerful AI. The wider the gap between the best-case and worst-case scenarios, the more severe the consequences of fluctuations become. Even if mitigation efforts are successful in maintaining an equilibrium on average, the increasing potential for damage means that the inevitable fluctuations will eventually lead to catastrophic outcomes. Given a long enough timeline, even a stable equilibrium will eventually be disrupted by a sufficiently large fluctuation, resulting in a catastrophic failure of mitigation.

So far we haven't argued that Mitigation will make the situation *worse*, only that there are significant Mitigation barriers. However, the idea that mitigation measures can be completely independent from prevention is implausible. In reality, mitigation is likely to have some impact on the probability and timing of AI failures, and this impact is problematic for two reasons.

First, mitigation may sometimes slow down and sometimes speed up disasters. When it slows them down, the increased damage potential of a more powerful AI leads to greater costs when failure eventually occurs. Imagine two scenarios: one with mitigation and one without. In the mitigation scenario, the time to failure is increased, but the damage at the point of failure is also higher. If the increase in damage is proportional to the increase in time to failure, then the total expected damage (calculated as the damage at the point of failure multiplied by the probability of failure) will be greater in the mitigation scenario.

Second, even if mitigation doesn't cause fluctuations in the average timing of failures, it can spread out the probability distribution of failure times. This means that failures become more likely to occur much earlier or much later than they would have otherwise. Given the steeply increasing damage potential of AI over time, the increased damage from later failures will outweigh the decreased damage from earlier failures. As an analogy, consider a game where you win $100 if a fair coin lands heads, but lose an amount equal to the square of the number of coin flips if it lands tails. Spreading out the probability distribution of the coin landing tails (by adding more flips) is a losing strategy, even if the average time to landing tails remains the same. For mitigation to be effective, it must not only avoid increasing the average time to failure, but also avoid spreading out the probability distribution of failure times.

Before leaving Mitigation behind, we should emphasize that some forms of Mitigation would promote measures that are both preventative and mitigating. To go back to our doomed climber: suppose we provided her with easy-to-grab-on-to handles attached to the mountain. The handles reduce the risk of a fall because they make climbing much easier and safer. Note that the handles also mitigate because when the climber falls she can grab onto the handles to slow the fall. This makes it harder to assess whether premise 2 in the non-deterministic argument is correct: the ranking in terms of height doesn't match the ranking in terms of badness. Climbing with handles could be a great idea: in expectation the severity of the fall's impact is less, even though the height is higher.

# 4. Conclusion and Looking Ahead

The non-deterministic argument presents a surprising and contrarian conclusion that challenges conventional assumptions about AI safety. Despite its counterintuitive nature, the argument proves remarkably robust and hard to respond to. For those who share our view that the non-deterministic argument presents challenge to standard views of AI safety, we propose several pathways for further research:

- **Develop additional responses:** While we have addressed several potential objections to the non-deterministic argument, there are many other strategies for reply that merit further exploration. For example, some may argue that creating super powerful AIs is impossible or that, even if possible, we will for some reason choose not to develop them. Or, maybe we can and would develop super powerful AI systems, but we won't because humanity will be exterminated before we ever get to that point. A very different line of reply, also worth exploring, rejects the underlying assumption that we can assign probabilities to these future events: the probability skeptic argues that given our limited understanding of the future trajectory of AI progress, the specifics of how an advanced AI might fail or what its consequences might be, assigning precise probabilities is a fraught exercise. One could also try to mix various responses (for example, a bit of credence in Optimism combined with some middling faith in some other responses.) Examining these and other counterarguments in greater depth could help to refine our understanding of the non-deterministic argument and its implications.

- **Challenge our responses to the replies**: Our analysis focused on Bottlenecking, Perfection Barrier, and Equilibrium Fluctuation. However, these responses may rely on empirical assumptions that require further investigation. Identifying and critically examining these assumptions could help to strengthen or modify our arguments.

- **Connect with ongoing AI safety research**: To bridge the gap between the abstract ideas presented in this paper and the practical challenges faced by AI safety researchers and developers, future work should explore how Optimism, Mitigation, Holism, Bottlenecking, Perfection Barrier, and Equilibrium Fluctuation manifest in specific AI systems and applications. By grounding these ideas in concrete case studies and empirical evidence, we can develop a more nuanced understanding of the practical implications of the arguments in this paper. What we have in mind here is relating the arguments here to the kinds of concrete efforts discussed e.g., in Amodei et al. (2016), Brundage et al. (2018), Christian (2020), Ord (2020), Russell (2019), and many other works on related issues. The main goal of this paper isn't to map these strategies onto current work on AI safety (we plan to do so in follow-up papers). Often it is unclear what the mapping is and that's unfortunate. If you're working on AI safety, and moved by the considerations in this paper, then that should be a pressing issue. Here are some potential connections: Optimistic approaches, such as value learning (Dewey 2011) and interpretable AI (Olah et al. 2018), aim to create inherently aligned and transparent systems. Mitigation strategies, like safe interruptibility (Orseau & Armstrong 2016) and containment (Babcock et al. 2019), focus on limiting the harm of failures. Adversarial robustness training (Goodfellow et al. 2014) and prompt injection attack resistance (Chen et al. 2024) illustrate the interplay between these approaches. Adversarial training mitigates damage from malicious exploits, but also optimistically seeks to proactively address vulnerabilities. One could think of prompt injection resistance as mitigating harm from manipulated inputs, while also holistically considering the broader context of deployment and potential misuse. In practice, these techniques often involve a mix of Optimism, Mitigation, and Holism.

- **Address implications for AI governance and policy**: If you share our sense that the non-deterministic argument is quite robust, what are the implications for AI governance and policy measures? Depending on how serious one takes the conclusion of our argument, the regulative consequences could include a ban on certain kinds of AI research, restrictions on hardware, or a redirection away from certain kinds of safety work. How all that should ideally play out is an open and complex question that we hope to explore in further work. Our goal here is to outline the core argumentative structure and an initial classification of response strategies[15].

---

[15] It is also worth elaborating on how the line of argument pursued in this paper differs from what we find in other writings on existential risk. That's a gigantic literature, so we'll just comment briefly on Nick Bostrom and Eliezer Yudkowsky. These writers share a concern about existential risk from AI, but they don't do so on the basis of the non-deterministic argument that, as we see it, captures the fundamental structure of the current situation. Bostrom's (2019) vulnerable world hypothesis argues that continued technological development will likely produce a technology capable of causing an existential catastrophe, but it does not explore the power-safety dynamic and the increasing risks that are central to our argument. Similarly, Yudkowsky's (2023) concern about AGI as a punctual event that will suddenly emerge and pose an immediate existential threat differs from our portrayal of AI development as an ongoing process with increasing risks. By not recognizing the non-deterministic nature of the situation and the interplay between AI power and safety, these views overlook the larger decision-theoretic puzzle that underlies our argument.

MacAskill, W. (2022). *What We Owe The Future*, Basic Books.

Maslej, N. et al (2023). *The AI Index 2023 Annual Report.* AI Index Steering Committee, Institute for Human-Centered AI, Stanford University. Available at: AI Index Report 2023 – Artificial Intelligence Index (stanford.edu)

Olah, C., Satyanarayan, A., Johnson, I., Carter, S., Schubert, L., Ye, K., & Mordvintsev, A. (2018). The building blocks of interpretability. *Distill*, 3(3), e10.

Ord, T. (2020). *The Precipice: Existential Risk and the Future of Humanity*. Bloomsbury.

Orseau, L. and Armstrong, S. (2016). Safely interruptible agents. In P*roceedings of the Thirty-Second Conference on Uncertainty in Artificial Intelligence* (pp. 557–566)*.* AUAI Press.

Russell, S. (2019). *Human Compatible: Artificial Intelligence and the Problem of Control*. OUP.

Solaiman, I., et al. (2019). Release Strategies and the Social Impacts of Language Models. *arXiv*. https://doi.org/10.48550/arXiv.1908.09203

Taylor, J., Yudkowsky, E., La Victorie, P. Critch, A. (2016). Alignment for Advanced Machine Learning Systems. in S. Matthew Liao (ed) Ethics of Artificial Intelligence (p. 342-382). 10.1093/oso/9780190905033.003.0013

Yudkowsky, E. (2008). Artificial intelligence as a positive and negative factor in global risk. In Bostrom. N. and Cirkovic, M. M. (Eds). *Global catastrophic risks* (pp. 308-345). OUP.

Yudkowsky, E. (2023). Pausing AI Developments Isn't Enough. We Need to Shut it All Down. *Time Magazine*, available at: https://time.com/6266923/ai-eliezer-yudkowsky-open-letter-not-enough20